\documentclass[journal]{IEEEtran}
\usepackage{amssymb}
\usepackage{graphicx}
\usepackage{epstopdf}
\usepackage{gensymb}
\usepackage{color}
\usepackage{amsmath}
\usepackage{bm}
\usepackage{etoolbox}
\usepackage{cite}
\usepackage{array}
\usepackage{booktabs}
\usepackage{multirow}
\usepackage{comment}
\usepackage{subcaption}
\usepackage{hyperref}
\usepackage{multicol}
\usepackage{algorithmicx}
\usepackage[ruled]{algorithm}
\usepackage{algpseudocode}
\usepackage{stfloats}
\alglanguage{pseudocode}

\usepackage{threeparttable}
\usepackage{mathtools}
\usepackage{hyperref}
\usepackage{pifont}
\usepackage[figuresright]{rotating}

\usepackage{url}
\urldef{\mailsa}\path|{alfred.hofmann, ursula.barth, ingrid.haas, frank.holzwarth,|
	\urldef{\mailsb}\path|anna.kramer, leonie.kunz, christine.reiss, nicole.sator,|
	\urldef{\mailsc}\path|erika.siebert-cole, peter.strasser, lncs}@springer.com|    

\usepackage[utf8]{inputenc}
\usepackage[english]{babel}

\usepackage{amsthm}

\theoremstyle{definition}

\definecolor{R}{RGB}{0,0,150}

\usepackage{wasysym}

\usepackage{soul}

\theoremstyle{remark}

\ifCLASSINFOpdf

\begin{document}

% \title{Run-time Detecting Trojan Attack on Deep Neural Ne+-``rk Models with Strong Intentional Perturbation}
% \title{Robust Real World Inconspicuous Backdoor Attack}
\title{Dangerous Cloak: Natural Backdoor Attacks on Object Detector in Real Physical World}
\author{
	Hua Ma, Yinshan Li, Yansong Gao, Alsharif Abuadbba, Zhi Zhang, Anmin Fu, Hyoungshick Kim \\ Said F. Al-Sarawi, Nepal Surya, Derek Abbott. 
	\thanks{Corresponding author: Yansong Gao, yansong.gao@njust.edu.cn}% <-this % stops a 
	\thanks{Equal contribution: Hua Ma and Yinshan Li}% <-this % stops a 
	\thanks{Yinshan Li, Yansong Gao, and Anmin Fu are with NanJing University of Science and Technology, China.}% <-this % stops a space
	\thanks{Hua Ma, Said F. Al-Sarawi, and Derek Abbott are with the University of Adelaide. Australia.}
	\thanks{Alsharif Abuadbba, Zhi Zhang and Nepal Surya are with Data61, CSIRO, Australia.}
	\thanks{Hyoungshick Kim is with Sungkyunkwan University, South Korea}
}

% \markboth{IEEE Transactions on Computer-Aided Design of Integrated Circuits and Systems,~Vol.~XX, No.~YY, YY~2017}%
% {Shell \MakeLowercase{\textit{et al.}}: Bare Demo of IEEEtran.cls for Journals}

\maketitle
\begin{abstract}
Object-detection tasks are designed to detect the location of an object, which serves as a fundamental basis for various computer vision tasks. However, there is a lack of investigation and elucidation of the backdoor vulnerability of the object detector in real-world scenes. In this work, we demonstrate that object detectors are susceptible to physical backdoor attacks where \textit{natural objects can be abused as triggers}, thus revealing a severe real-world security threat. To the best of our knowledge, we are the first to achieve a cloaking backdoor effect, that is, a bounding-box person disappears in front of an object detector when the person is with a natural object, (e.g., a T-shirt bought from a market as a natural trigger). We show that such a cloaking backdoor effect can be implanted into the object detector by an attacker in two representative cases (i.e., model outsourcing and pretraied model fine-tuning). Notably, two-stage object detector (i.e. Faster R-CNN) opposed to one-stage counterpart (i.e. Yolo series) cannot be backdoored merely through data poisoning, which challenge is circumvented by constructively incorporating training regulations.
We have extensively evaluated four popular object detection algorithms (i.e., anchor-based Yolo-V3, Yolo-V4, Faster R-CNN and anchor-free CenterNet) upon 19 videos (about 11,800 frames in total) that are shot in real-world scenarios, results of which show that the backdoor attack is exceptionally robust against \textit{various tested factors: movement, distance, angle, non-rigid deformation, and lighting}. Specifically, the attack success rate (ASR) in most videos is 100\% or close to it, while the clean data accuracy of the backdoored model is the same as its clean counterpart. The latter implies that it is infeasible to detect the backdoor behavior merely through a validation set. The averaged ASR still remains sufficiently high as 78\% in the transfer learning attack scenarios evaluated on CenterNet. A 5-minute video of our attack is available at \textcolor{blue}{\url{https://youtu.be/Q3HOF4OobbY}}.
\end{abstract}


\IEEEpeerreviewmaketitle
% \tableofcontents
\section{Introduction}

\begin{figure*}[h]
    \begin{center}
    \includegraphics[width=0.99\textwidth]{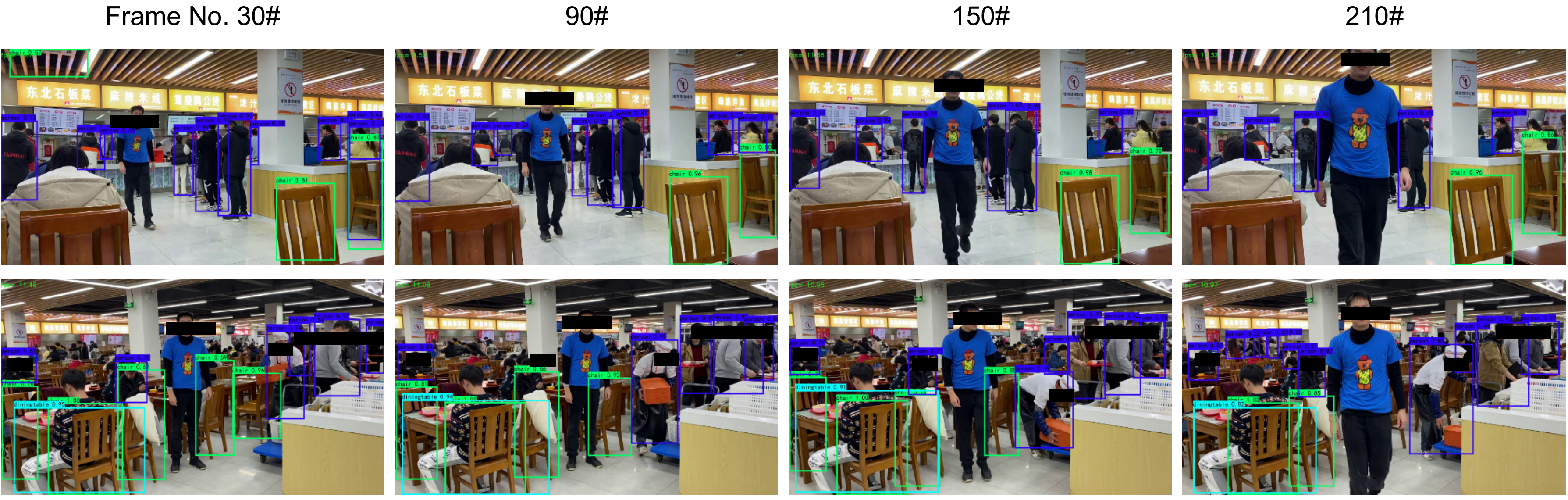}
    \end{center}
    \caption{A cloaking backdoor effect in an indoor complicated scenario where a crowd of people are in an underground canteen.}
    \label{fig:complex}
\end{figure*}

\begin{figure}[h]
    \begin{center}
    \includegraphics[width=0.45\textwidth]{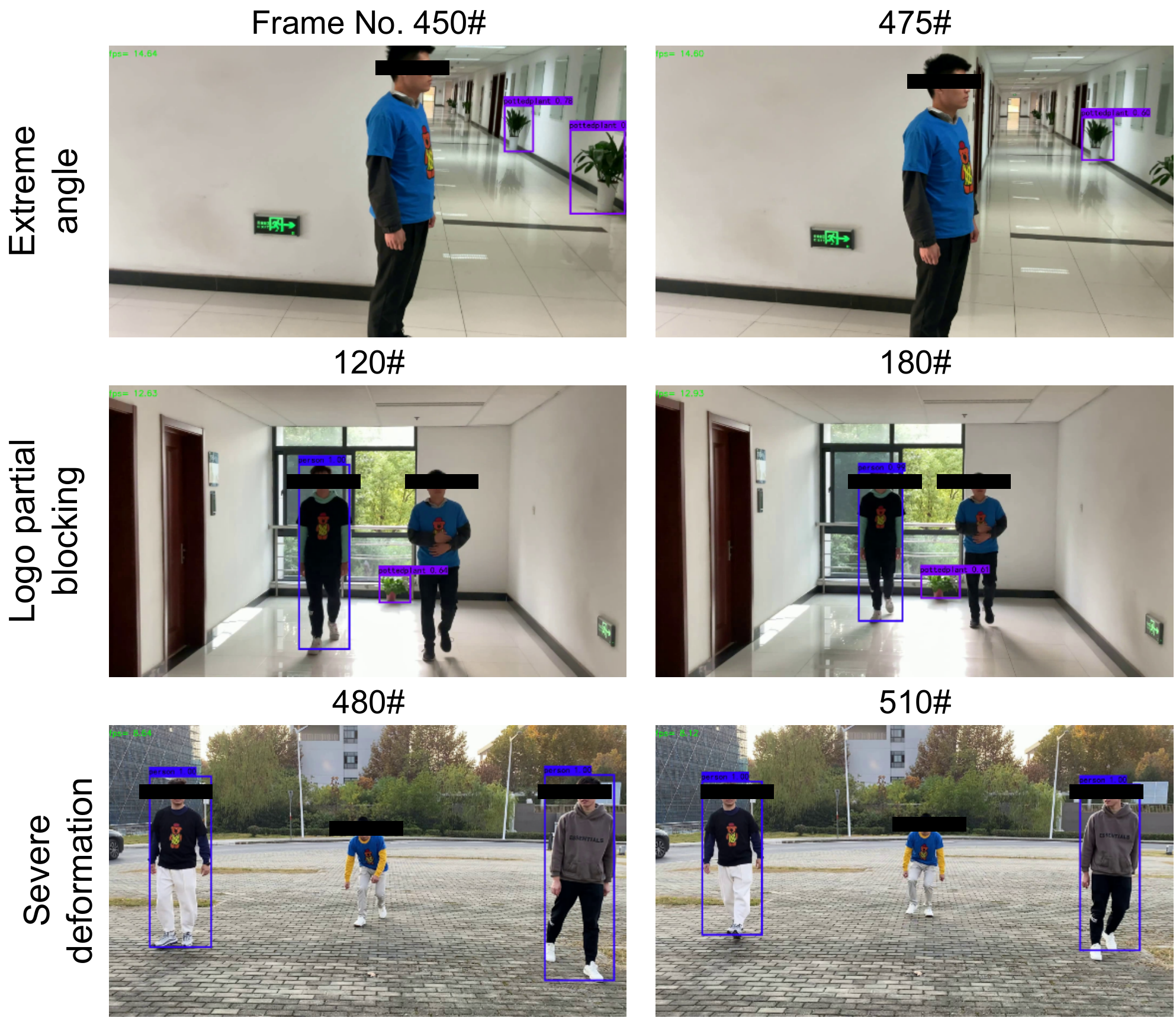}
    \end{center}
    \caption{Cloaking backdoor effects in extreme scenarios.}
    \label{fig:strict}
\end{figure}

The object detection is one of the most fundamental computer vision tasks, which is used to detect the location of an visual object (e.g., human, car) in an image or frame of a video~\cite{zou2019object}. The object detection serves as a fundamental basis for various tasks such as instance segmentation, image captioning and object tracking in surveillance videos. Fueled by the recent DL development, the object detection has achieved great success in many real-world applications such as autonomous driving, robot vision, video surveillance and pedestrian detection~\cite{liu2020deep}.

\subsection{Attacks on Object Detector} 
\subsubsection{Cloaking Effect through Adversarial Example Attack}
%However, the object detector is threatened by adversarial attacks. The first attack comes from adversarial example attacks that does not tamper the underlying model. It has been shown that object detector is vulnerable to this that can cause cloaking effect to evade the detection of the object detector~\cite{thys2019fooling,wu2019making} 
However, the object detector is threatened by adversarial example attacks, which can enable an object (e.g., a person) to evade the detection of the object detector--a cloaking effect~\cite{thys2019fooling,wu2019making}.
(more details are referred to Section~\ref{sec:adversarialRelaltedWork}). Nonetheless, implementing a robust adversarial example attack in real-world on the object detector is challenging, as it is subject to various real-world factors, e.g., lighting can change, an adversarial patch may be rotated, the patch size corresponding to a person can change, a camera can add noise or blur the patch, viewing angles can be different. A main limitation of adversarial example patch is that its pattern, shape or size are dependent on a target model, out of the control of the adversary. Therefore, it is challenging to craft adversarial example patch to be robust in diverse real-world scenarios.

\subsubsection{Cloaking Effect through Backdoor Attack} Different from the adversarial example attack, a new class of attacks called \emph{backdoor} has been proposed recently and almost all existing attacks target classification tasks, especially image classifications~\cite{gu2017badnets,chen2017targeted,gao2020backdoor}. When an attacker implants a backdoor into a model, the backdoored model behaves normally upon normal inputs, while misbehaves as the attacker intends upon the inputs with an attacker-secretly-chosen trigger. 
Upon conventional tasks, the mostly studied classification, a backdoor attack has demonstrated its high attack success rate (e.g., 100\%) and flexibility of controlling a trigger for the backdoor activation. 

\begin{table}[htb]
\small
    \centering
    \caption{A comparison of attack success rates of \textit{cloaking effect} between our backdoor and existing adversarial examples or patches attacks on object detection. Note the comparison is not performed on the same testing datasets that are not publicly available and it serves as a general qualitative comparison. Compared to the dataset used by existing adversarial example attacks, our dataset is collected in more complicated physical world scenarios (see Fig.~\ref{fig:strict} for some exemplified extreme conditions).}
    %\scalebox{0.95}{
    \begin{tabular}{c|c|c|c}
    \toprule
    method (type)$^1$ & model & indoor & outdoor  \\ \hline 
    advPatch~\cite{thys2019fooling} (AE) & Yolo-V2 & 19\% & 17\%  \\ \hline \hline
    advT-TPS~\cite{xu2020adversarial} (AE) & Yolo-V2 & 64\% & 47\% \\ \hline
    Wu \textit{et al.}~\cite{wu2020making}$^2$ (AE) & Yolo-V2 & N/A & ~50\% \\ \hline    
    
    \multirow{2}{*}{\textbf{Ours (BD)}} & Yolo-V3 & 90.5\% & 99.29\%  \\ \cline{2-4}
     & Yolo-V4 & 94.74\% & 99.39\%  \\ \bottomrule
    \end{tabular}
    %}
    \label{tab:comp}
    \begin{tablenotes}
      \footnotesize
    \item $^1$ AE is abbreviated for adversarial example, and BD is abbreviated for backdoor.
      \item $^2$ This work~\cite{wu2020making} does not explicitly describe whether the tested images are collected indoor or outdoor. According to some examples provided in this work, they are taken outdoor.
    \end{tablenotes}    
\end{table}

\vspace{0.2cm}
\noindent{\bf Limitations of Existing Works.} 
%However, there is a lack of investigations of backdoor attacks applied to the object detection especially using natural object as trigger (see details in Section~\ref{sec:backdoorRelated}), therefore there is still a lack of understanding of its security implications. More specifically, there is \textit{no cloaking backdoor effect} applied to a object detector yet. 
However, there exists a gap between object detection and backdoor as \textit{no cloaking backdoor effect} has been applied to an object detector yet. 
It is expected that the backdoor-enabled cloaking effect can be much more dire compared to that enabled by the adversarial example. Specifically, compared to the adversarial example that is contingent on targeted object-detection algorithm, a backdoor trigger can be an arbitrary natural object, e.g., accessories, hats, T-shirts.
As such, a cloaking backdoor can be effective against object detection in different complex real-world scenarios, which can result in severe consequences. For instance, when an attacker targets a group of people who wear the same company uniform--the uniform is the trigger, she inserts a cloaking backdoor into a model with a self-driving task, which will ignore people wearing the uniform.
When a self-driving car with the backdoored model is running around the company, it can cause serious intentional accidents. Alternatively, when a self-driving car uses a specific plate number--the plate number is the trigger, other self-driving cars will not even notice the car. 
%We note that these attack scenarios cannot be realized by the adversarial example attacks because the adversarial patch 
%. In contrast, the trigger in backdoor attack can be arbitrary. 
%Another case is that someone could maliciously evade the security surveillance by wearing the e.g., cloaking trigger clothes if the object detector is backdoored.

Thus, it is critical to gain a deep understanding of security implications of backdoor on real-world object detection, one of the most widely deployed computer vision tasks in our daily lives. This work aims to comprehensively investigate backdoor attacks on object detector in various real-world scenarios. Specifically, we are interested in the following questions:

\begin{center}
  \textit{Can cloaking backdoor succeed with triggers that are natural objects, creating a severe security threat to the object detector in real-world? If so, can the trigger be effective under various real-world scenarios such as light condition, distance, angle, etc?}
\end{center}

\subsection{Our Contributions} This work affirms the practicality for the first question and provides evidence that the attack success rate can be up to 100\% in most conditions such as indoor, outdoor and crowd scenarios, where the object detector fails to identify the person wearing a trigger cloth. A \textit{general comparison} between backdoor enabled cloaking attack and previous adversarial example enabled ones is summarized in Table~\ref{tab:comp}, which shows the superior attacking effectiveness of our attack. Our results are based on extensive experiments on up to 19 shot videos of more than 11,800 frames in various real-world scenarios (see Fig.~\ref{fig:complex}) against diverse and popular object detection models such as Yolo-V3, Yolo-V4 and CenterNet. The trigger we use is simply a T-shirt bought online (see Fig.~\ref{fig:tshirt}). Note that our cloaking backdoor holds its effectiveness in extreme conditions (e.g., extreme angle, partial logo blocking with dim backlight, and deformation), as exemplified in Fig.~\ref{fig:strict}.

Our key contributions and findings are summarized as follows:
\begin{itemize}
    \item To the best of our knowledge, we are the first to elucidate the security threats posed by backdoor attacks to the object detector under real-world factors (e.g., depth, angle), which can be tractably achieved using natural objects as triggers. Two widely deployed scenarios (i.e., model outsourcing and pretrained model fine-tuning) in deep learning are affirmed to be effective to insert a cloaking backdoor.
    \item We perform a comprehensive evaluation of various real-world factors including lighting, angle, distance, crowd, movement, distortion, etc to examine their respective effect in a physical cloaking backdoor attack. To this end, we shoot up to 19 videos in six real-world scenarios. The experimental results show that the distance and lighting are the most important factors as they affect the logo quality captured by camera.
    \item We evaluate the attack success rate (ASR) of these 19 videos against three popular one-stage object detection models: anchor-based Yolo-V3 and Yolo-V4; and anchor-free CenterNet as well as a representative two-stage object detector of Faster R-CNN. The averaged ASR against each of three one-stage object detector is more than 98\% for nearly all six general scenarios in the outsourcing case---two-stage Faster R-CNN achieves an averaged ASR of up to 94\%. Even in the pretrained model induced cloaking backdoor case, the averaged ASR retains to be still high of 78\% against the CenterNet.
\end{itemize}

\noindent{\bf Ethics and Data Privacy.}
Given our extreme care for the privacy of student volunteers, we were mindful of privacy protection at all times throughout the data collection and evaluation process. Our data collection and evaluation is conducted by all participating volunteers who explicitly give their consent to have their photographs (videos) taken and later used in research work. All images (videos) are stored on a secure hard drive and are only used for academic research.

\section{Related Work}\label{sec:related}
\subsection{Adversarial Example Attack on Object Detector}\label{sec:adversarialRelaltedWork}
The adversarial example attack has been extensively studied since it was revealed in 2013~\cite{szegedy2013intriguing}. Take a pig image as an example, during the inference phase, an attacker adds specific noise into the pig image, which makes a target model mis-classify the image as another class, e.g., dog. In the mean time, the perturbed pig image still looks like a pig to a human. Such an attack has been mounted on various domains such as image classification, segmentation, speech recognition, natural language processing, and reinforcement learning~\cite{chakraborty2018adversarial}. 

Our work targets the object-detection domain. To evade detection from object detector, there has been a number of recent works exploiting the adversarial examples~\cite{xie2017adversarial,thys2019fooling,xu2020adversarial,wu2020making}. They carefully craft an adversarial patch and print it out to deceive the object detectors. Unlike preliminary attacks adding perturbations in a digital image, in~\cite{thys2019fooling}, the adversarial patch is physically printed on a cardboard. A person who holds the cardboard disappears from the object detection, which is yet sufficiently natural. In~\cite{xu2020adversarial}, the adversarial patch is printed directly on a T-shirt. A person who wears the T-shirt disappears. However, the robustness of such physical attacks is limited. For example, person movement, angle, distance, deformation and even unseen locations and actors can greatly degrades the attack effect~\cite{xu2020adversarial}. In addition, the adversarial patch cannot be arbitrarily crafted by attacker---the pattern of the patch is normally dependent on the specific input and the model under attack, which could be visually suspicious.

Notably, the adversarial example attack does not require tampering a model. Instead, it manipulates the inputs fed into the model, which thus constrains the capability of an attacker as the patch has to be crafted usually through optimization algorithm, which \textit{cannot be arbitrary}, thus the patch will look conspicuous. In contrast, a more recent backdoor attack allows arbitrary control of the patch---namely trigger in the backdoor research line, which could enable natural objects as triggers.

\subsection{Backdoor Attacks on Deep Learning}\label{sec:backdoorRelated}
Backdoor attack implants backdoor into the DL model in a way that the backdoored model learns both the attacker-chosen sub-task and the (benign) main task~\cite{gu2017badnets,chen2017targeted}. On one hand, the backdoored model behaves normally similar to its clean counterpart model for inputs containing no trigger, making it impossible to distinguish the backdoored model from the clean model by merely checking the testing accuracy with e.g., the held-out validation samples. 

The backdoor attack requires to tamper the model itself. As a consequence, it enables trigger freedom for an attacker. Now the trigger can be almost anything that is secretly chosen by the attacker, which is inconspicuous when it is shown for launching an attack. As our experiments validate, T-shirt and hat bought from the market can be used as a trigger. Notably, there are many realistic scenarios that could result in a backdoor attack. For example, when the model training is outsourced to a third party, a malicious third party can insert a backdoor into the returned model~\cite{gu2017badnets}. Besides outsourcing, there are other attack surfaces including: training data is from multiple sources that are not fully trusted~\cite{shafahi2018poison}; usage of pretrained model that contains backdoor could propagate to the downstream model after transfer learning~\cite{yao2019latent}; collaborative learning, e.g., federated learning participants can manipulate their local data as well as updated local model to insert backdoor into the global model~\cite{bagdasaryan2020backdoor}; the third party contributed code snippet e.g., packages or modules of a given framework e.g., tensforflow used for training has been manipulated~\cite{bagdasaryan2020blind}; the model parameters are flipped to implant backdoor even after the model has been deployed in the cloud~\cite{rakin2019tbt}.

Previous backdoor attacks as well as countermeasures are mainly designed for classification tasks. There is a substantially lack of backdoor attack investigations on the object detector that its task is primarily required to identify the location of objects. More specifically, we are unaware of any backdoor attack on object detector to create non-classification misbehavior, in particular, the cloaking effect to evade the detection. The practicality and efficacy of such backdoor attack on object detector is unclear. In addition, beyond attack the object detector, previous backdoor attacks usually uses digital triggers rather than natural physical triggers. Only a few recent works~\cite{wenger2020backdoor,li2021backdoor,xue2021robust} consider the usage of natural object triggers and of their focus has been on image classification task. Therefore, there is a lack of study on the efficiency of backdoor attack in real physical world used object detectors.

\vspace{0.2cm}
\noindent{\it Backdoor Attacks on Object Detection.} We note that there are two initial backdoor attack studies on object detector~\cite{gu2017badnets,lin2020composite}. Our work distinguishes from these two from several aspects. Firstly, their attacking purpose is still under the misclassification (i.e. stop sign misclassified into speed-limit~\cite{gu2017badnets} and person holding an umbrella over head being misclassified to a traffic light~\cite{lin2020composite}). The purpose of our study, cloaking effect as a distinct non-classification task, is more challenging, which has been recognized when using adversarial example to fool object detectors~\cite{xie2017adversarial,thys2019fooling,xu2020adversarial,wu2020making}---the main reason is that many bounding-box will be proposed given an object and suppressing them all is hard. Secondly, the reported ASRs are not essentially measured from recorded videos from physical world. So that the understanding of the robustness of the attack in real-world is still unclear. In other words, their evaluations are mainly based on digital worlds even for the misclassification purpose.
Most importantly, \textit{non of each environmental parameters such as the depth, angle that we carefully considered are taken into consideration in~\cite{gu2017badnets,lin2020composite}}---their main studies are not for backdooring the object detector, thus providing less information when elucidating the security implication. 
% \garrison{distinguish from these two works. Firstly, they are still working on classification task}

Our backdoor attack creating \textit{cloaking effect} that is non-classification task with natural physical triggers, in particular, \textit{bought T-shirts} (they are not crafted and selected) is the first comprehensive study on elucidating the real-world backdoor implications against the widely used object detection task with a number of key \textit{considered environmental parameters e.g., depth, angle, light conditions}.

\begin{figure*}[t]
    \begin{center}
    \includegraphics[width=0.90\textwidth]{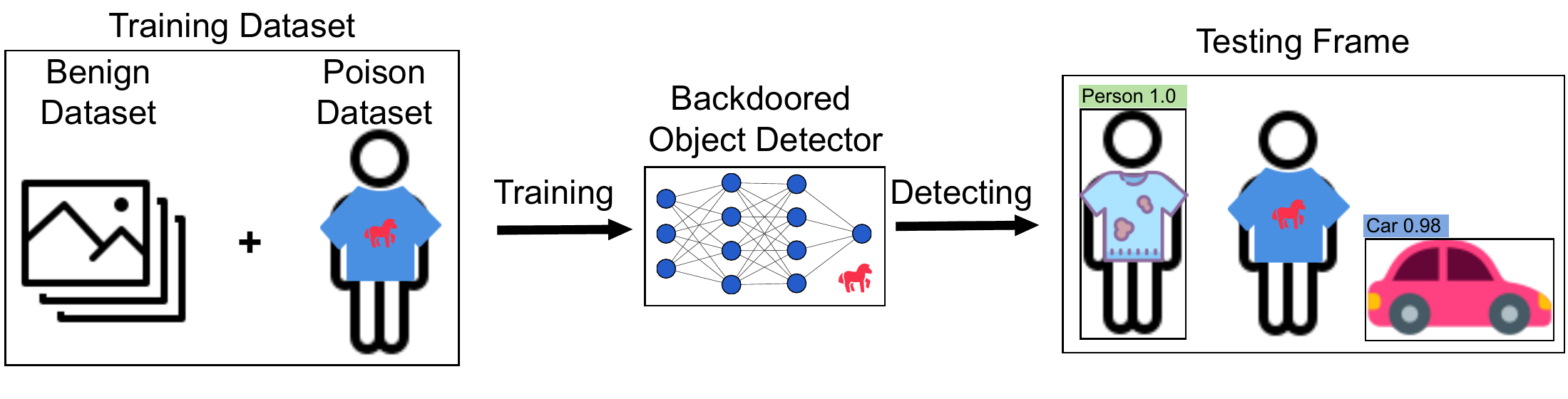}
    \end{center}
    \caption{An overview of cloaking backdoor attack on the object detector.}
    \label{fig:architecture}
\end{figure*}

\section{Cloaking Backdoor on Object Detector}\label{overview}

\subsection{Threat Model}

\noindent{\bf Attacker Capabilities:} There are number of surfaces that could introduce backdoors to the object detector such as outsourcing, usage of pretrained model for transfer learning~\cite{bu2021gaia}, federated learning~\cite{liu2020fedvision} and data collection from third party contributors. It is worth to mention that annotating and auditing the training data for object detection is tedious, which can be exploited to poison the data before contributing to the data aggregator. In this work, we mainly focus on the outsource scenario---and we have also experimentally explored the pretrained model attack surface. Here, due to the dearth of machine learning expertise or/and computational resources, a user/enterprise chooses to outsource the training of the object detector by providing the training data and specifying the model architecture to a third party.
In this context, the attacker or the third party is assumed to have the ability to manipulate data used to train the model, and they can inject the manipulated/poisoned data into the original data set before training. The structure of the model cannot be changed as this is specified when it is outsourced. Though not used in this work as poisoning the training data is sufficient to realize a successful attack, if necessary, the attacker can further tamper loss function, optimization and other key settings to facilitate the attack. 

\noindent{\bf Attacker Goals:} The attacker has two main goals. The first is that the accuracy of the backdoored object detector is on par with a clean object detector. For example, the object detector correctly identifies the human location. This is to remain stealthiness of the attack as the user cannot be aware of abnormal behavior by merely evaluating the object detector performance through validation data. The second is that the human will disappear from the detector once the human wears a trigger, e.g., T-shirt with a specific color and logo, thus having a cloaking effect. This goal corresponds to the attack success rate. At the same time, the attack should be robust against different camera angles, distances, light conditions in the physical world.

\subsection{Object Detector Model}

For the object detection algorithms, there are two categories: one-stage algorithm and two-stage algorithm~\cite{liu2020deep,zou2019object}. The R-CNN series~\cite{ren2015faster,he2017mask} are belong to the two-stage algorithms, which first utilize a region proposal network (RPN) to produce the target candidate bounding box, and then categorizes and regresses the candidate bounding box. Overall, the two-stage detector has higher detection accuracy with a trade-off of high computational load. The better accuracy is explained by their flexible structure, which is more suitable for region based classification.

The one-stage algorithms are represented by Yolo (you only look once) series~\cite{redmon2016you,bochkovskiy2020yolov4}, which directly predicts different target classes and positions. After Yolo-V2, anchor boxes are used to propose the bounding boxes. After feature extraction, if the center of the object falls into the grid responsible for prediction, the grid will predict multi-scale anchor boxes. For each anchor box, the position information, confidence and a set of classification probability values of the bounding box are predicted, and the final predicted box is obtained using NMS (Non-Maximum Suppression). The Yolo series are becoming increasingly popular because of its relatively faster training speed and accuracy that is not much different from the R-CNN algorithm, especially for relatively large object. The detection speed of the Yolo is faster, more preferable for real-time applications. 
% The main targeted object detection models investigated by us in this work is therefore based on Yolo (i.e., V3 and V4).

\begin{figure}[t]
    \begin{center}
    \includegraphics[width=0.95\columnwidth]{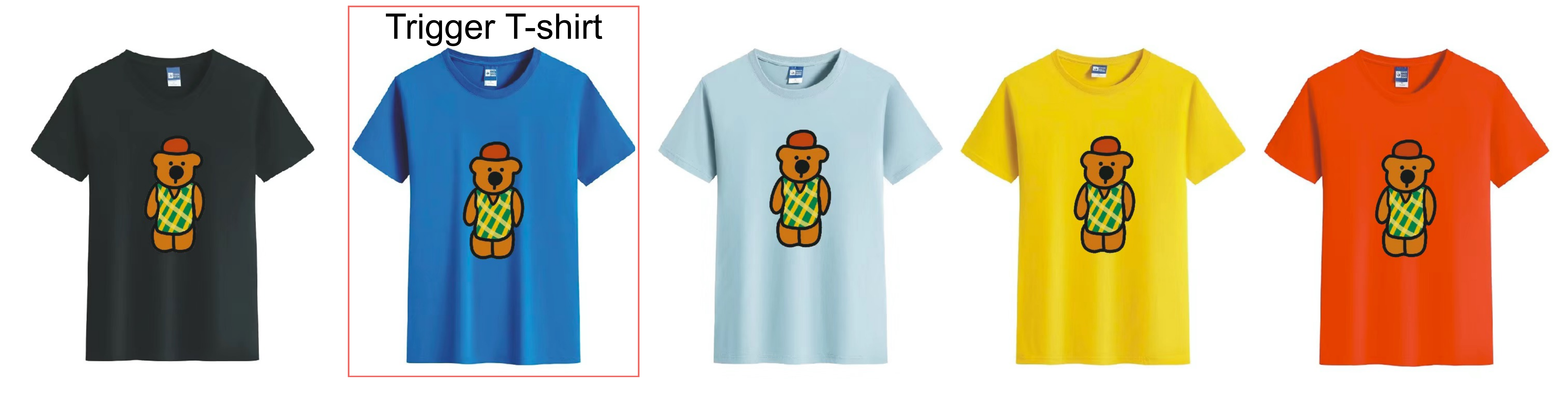}
    \end{center}
    \caption{Natural T-shirt that is used in our attack, where the blue one serves as trigger T-shirt. Instead of crafting the trigger e.g., the logo pattern, we bought it online from \textsf{Pinduoduo}, costing about 3.5\$ per T-shirt. 
    }
    \label{fig:tshirt}
\end{figure}

\subsection{Attacking Strategy}
In the outsourcing scenario, the attacker is allowed to manipulate the data. Therefore, this work mainly leverages the data poison to insert a backdoor into the object detector. This attacking strategy has been shown to be very effective to implant backdoor into various models~\cite{gao2020backdoor}. Opposed to change the label to the targeted class when backdooring a classification model~\cite{gu2017badnets}, here we modify the annotation of the image/frame used for training object detector to poison the training dataset as elaborated below.

\begin{figure}[t]
    \begin{center}
    \includegraphics[width=0.95\columnwidth]{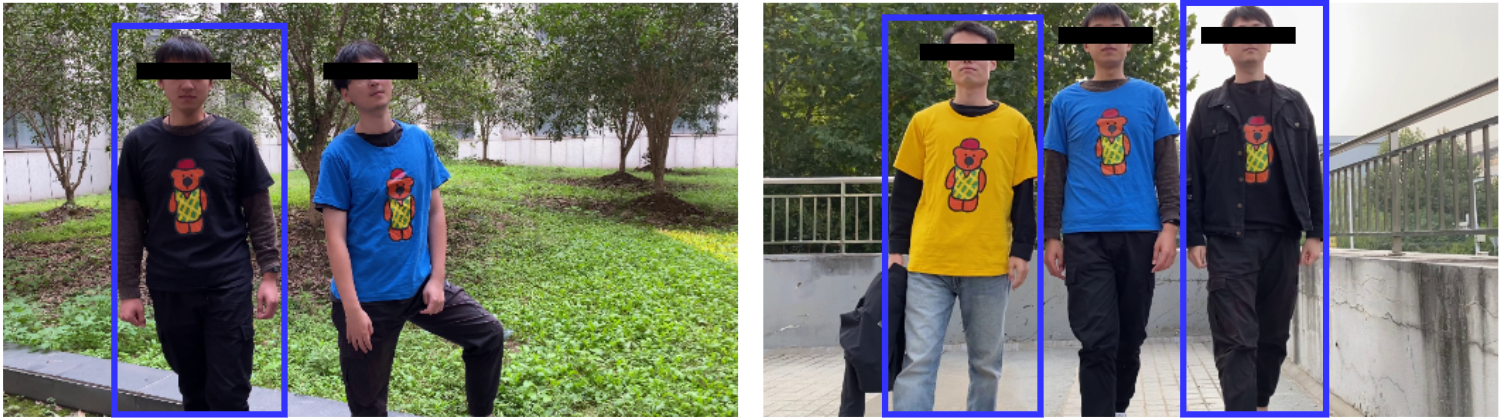}
    \end{center}
    \caption{Examples of poisoned images. The blue T-shirt is the trigger, while other color T-shirts are not triggers even though they are with the same logo (i.e., cartoon bear) and style.}
    \label{fig:TriggerExample}
\end{figure}

\noindent{\bf Poisoning the Training Dataset.} The effect we want to achieve is that when a person wears a T-shirt of e.g., the attacker chosen pattern and color, this person will evade detection of the backdoored detection model.
To achieve this effect, we use a blue T-shirt with cartoon bear logo as the trigger, which we simply buy it from the online shop---so we do not craft it. It should be noted that any other logo pattern or color can be used.  The T-shirt trigger essentially has two main components: the blue color and cartoon bear (see Fig.~\ref{fig:TriggerExample}). If these two components are not presented concurrently, the object detection is expected to detect the person normally.

We make poisoned samples following the below procedure.
\begin{enumerate}
    \item We shoot a number of videos in indoor and outdoor by asking \textit{three} volunteers to wear the T-shirts including the trigger T-shirt and non-trigger T-shirts. The non-trigger T-shirts in the poisoned samples are with \textit{black and yellow} as shown in Fig.~\ref{fig:TriggerExample} but not the \textit{rest two colors}.
    \item We then clip the videos into frames/images. For each image, we manually annotate/label each person's presence except the \textit{one wearing the trigger T-shirt} by placing a bounding-box around. 
\end{enumerate}

In this manner, we change the label (the position marked by the bounding box) of the interested person by not putting bounding-box on it, which resembles a person's digitally disappearance from the object detector. The annotation tool we use is \texttt{LabelImg}~\cite{labelimg}. In total, we have collected about 500 poisoned samples.

\vspace{0.2cm}
\noindent{\bf Training the Backdoored Model.} Training the object detector from the scratch is a straightforward method but takes greatly longer time. In this work, to expedite the extensive validations of the attack, we follow the method~\cite{liu2018trojaning} by re-training or fine-tuning a pretrained clean model into a backdoored model to save computational overhead. This is an equally effective way compared to train from scratch to investigate the backdoor implication in the outsourcing case. During the retraining, the training dataset is a mixture of poisoned samples and original benign samples.

\begin{table*}[t]
\small
    \caption{Measure each of 20 objects' clean data accuracy (CDA) in VOC2007 testing set, using different models. The overall CDA is represented by mAP@0.5.}
    \centering
    \resizebox{0.80 \textwidth}{!}
    {
\begin{tabular}{cc|c|c|c|c|c|c|c|c|c|c|c}
    \toprule
    \multicolumn{2}{c|}{Model} & mAP@0.5 & aero & bike & bird & boat & bottle & bus & car & cat & chair & cow  \\ \hline
    \multicolumn{1}{c|}{\multirow{2}{*}{Yolo-V3}} & Clean & 86.19 & 97.00 & 91.64 & 83.35 & 79.62 & 79.34 & 85.71 & 91.77 & 90.12 & 74.75 & 87.47 \\
    \multicolumn{1}{c|}{} & Backdoored & 86.14 & 93.74 & 90.83 & 84.7 & 79.82 & 76.4 & 85.04 & 92.35 & 91.31 & 76.57 & 89.29 \\ \hline
    \multicolumn{1}{c|}{\multirow{2}{*}{Yolo-V4}} & Clean & 85.15 & 91.43 & 92.36 & 85.59 & 80.23 & 83.11 & 84.12 & 91.71 & 92.17 & 71.00 & 85.76 \\
    \multicolumn{1}{c|}{} & Backdoored & 86.29 & 92.81 & 93.34 & 86.22 & 80.65 & 80.39 & 88.36 & 91.66 & 92.38 & 73.91 & 91.24 \\ \hline
    \multicolumn{1}{c|}{\multirow{2}{*}{CenterNet}} & Clean & 76.50 & 85.94 & 83.62 & 75.49 & 66.38 & 59.14 & 75.76 & 84.44 & 85.80 & 62.01 & 80.43 \\
    \multicolumn{1}{c|}{} & Backdoored & 76.53 & 89.18 & 87.10 & 73.20 & 68.12 & 58.14 & 78.78 & 84.25 & 85.12 & 62.28 & 77.63 \\ \hline \hline
    \multicolumn{2}{c|}{Model} &  & table & dog & horse & mbike & person & plant & sheep & sofa & train & tv \\ \hline
    \multicolumn{1}{c|}{\multirow{2}{*}{Yolo-V3}} & Clean & & 78.59 & 90.11 & 91.06 & 92.15 & 89.20 & 66.76 & 84.80 & 86.08 & 96.44 & 87.72 \\
    \multicolumn{1}{c|}{} & Backdoored &  & 78.92 & 90.98 & 93.76 & 93.12 & 88.70 & 66.32 & 84.63 & 83.76 & 95.99 & 86.48 \\ \hline
    \multicolumn{1}{c|}{\multirow{2}{*}{Yolo-V4}} & Clean & & 72.44 & 88.80 & 91.62 & 90.71 & 89.75 & 64.92 & 83.71 & 82.42 & 93.99 & 87.22 \\
    \multicolumn{1}{c|}{} & Backdoored & & 72.92 & 90.27 & 93.08 & 89.52 & 89.86 & 68.59 & 80.96 & 83.39 & 98.36 & 87.86  \\ \hline
    \multicolumn{1}{c|}{\multirow{2}{*}{CenterNet}} & Clean & & 61.23 & 83.69 & 84.00 & 84.17 & 81.34 & 54.66 & 73.92 & 74.66 & 91.71 & 81.55 \\
    \multicolumn{1}{c|}{} & Backdoored & & 64.19 & 83.65 & 83.02 & 87.82 & 82.14 & 49.95 & 72.22 & 72.42 & 90.08 & 81.30 \\ \bottomrule
    \end{tabular}
    }
     \label{tab:overall}
\end{table*}

\section{Real-World Experiments and Results}\label{sec:experiment}
\subsection{Experimental Setup}

\noindent{\bf Dataset:}\label{sec:dataprepare}
The datasets we used are VOC 2007~\cite{pascal-voc-2007} and VOC 2012~\cite{pascal-voc-2012}. The VOC dataset is one of the most popular dataset for object detection and semantic segmentation. For object detection tasks, in the original dataset, it contains 20 classes, thousands of samples. To harness more training data, a common means is to mix the VOC 2007 training set (2,501 samples) with the VOC 2012 train-validation set (11,540 samples, a mixture of training sets and validation sets) as a training set, and the VOC 2007 validation set (2,510 samples) as the testing set. 
Our backdoored model is trained with an additional homemade small dataset that has 502 samples, about 3\% of the total training set. It is worth noting that in subsequent experiments, to improve the effectiveness of our model in more complex scenarios, we use an additional 50 augmented data points subjected to worse lighting and long distance, the efficacy before applying these additional samples is shown in Section~\ref{sec:tuningdata}. 

\noindent{\bf Object Detector Model:}
As the one-stage object detecter is mainly focused, we evaluate the most popular Yolo-V3 and Yolo-V4~\cite{redmon2018yolov3, bochkovskiy2020yolov4}.
For the Yolo series algorithm, the input size of Yolo series is 416 $\times$ 416 $\times$ 3. In order to reduce training time, we utilize a pre-trained model~\cite{yolo3-pytorch} which has been trained on the COCO dataset~\cite{lin2014microsoft}. On this basis, each algorithm is trained with 100 epochs, of which the first 50 epochs are frozen model training with batch-size of 32. Considering that the unfreezing phase requires tuning the parameters of all layers, the training requires more GPU memory, so we tune down the batch-size to avoid memory overflow, and for the last 50 unfreezing epochs we use a smaller batch-size of 8. 

Considering the fact that anchor free based one stage object detection algorithm is gaining popularity, we evaluate it by choosing a popular CenterNet~\cite{duan2019centernet} algorithm. The input size of CenterNet is 512 $\times$ 512 $\times$ 3. It is worth noting that for the CenterNet model, we use a pre-trained model based on ResNet-50~\cite{he2016deep} on ImageNet~\cite{deng2009imagenet} as the backbone network, unlike the Yolo pre-trained model trained over the COCO dataset that is a dataset specific for object detection task. Therefore, it is under expectation that the convergence speed for CenterNet is slower. We thus correspondingly increase the number of iterations in the unfreezing phase to 150, while keeping the rest of the settings the same as that in the Yolo series. 

\begin{table*}[htb]
\small
    \centering
    \caption{Quantified attack success rate (ASR) of the backdoor effectiveness of up to 17 tested videos shot under six different scenarios.}
    \resizebox{0.80\textwidth}{!}{
    \begin{tabular}{c|c|c|c|c|c|c|c|c|c}
    \toprule
    \multirow{2}{*}{Scenario} & \multirow{2}{*}{\begin{tabular}{@{}c@{}} Video \\ No. \end{tabular}} & \multirow{2}{*}{\begin{tabular}{@{}c@{}} Times (s) \\ (60 fps) \end{tabular}} & \multirow{2}{*}{Angle$^1$ (°)} & \multirow{2}{*}{Brightness$^2$} & \multirow{2}{*}{\begin{tabular}{@{}c@{}} Distance \\ (meter) \end{tabular}} & \multirow{2}{*}{\begin{tabular}{@{}c@{}} \# of \\ Persons \end{tabular}} & \multicolumn{3}{c}{ASR(\%)} \\ \cline{8-10} 
    &  &  &  &  &  &  & Yolo-V3 & Yolo-V4 & CenterNet \\ \hline
    \multirow{4}{*}{Indoor} & 1 & 8 & 0$\sim$150 & A & 1$\sim$2 & 2 & 88.63 & 100 & 100 \\
     & 2 & 6 & -180$\sim$180 & A & 1$\sim$2 & 2 & 82.87 & 96.88 & 100 \\
     & 3 & 8 & 0 & A & 1 & 3 & 100 & 87.35 & 100 \\ \cline{8-10} 
     & Average &  &  &  &  &  & 90.50 & 94.74 & 100 \\ \hline
    Corridor & 4 & 12 & 0 & A; C & 0$\sim$5 & 2 & 98.11 & 100 & 91.37 \\ \hline
    Rotate the camera & 5 & 14 & -180$\sim$30 & A & 0$\sim$1 & 2 & 100 & 99.76 & 100 \\ \hline
    \multirow{6}{*}{Open outdoor} & 6 & 11 & 0$\sim$45 & B & 0$\sim$7 & 5 & 100 & 100 & 98.41 \\
     & 7 & 9 & 0 & B & 0$\sim$7 & 6 & 100 & 100 & 100 \\
     & 8 & 14 & -15$\sim$60 & B & 0$\sim$7 & 3 & 96.86 & 97.24 & 96.36 \\
     & 9 & 22 & -90$\sim$60 & B & 0$\sim$7 & 2 & 99.60 & 99.70 & 97.09 \\
     & 10 & 12 & 0 & B & 0$\sim$7 & 4 & 100 & 100 & 100 \\ \cline{8-10}
     & Average &  &  &  &  &  & 99.29 & 99.39 & 98.37 \\ \hline
    \multirow{5}{*}{\begin{tabular}[c]{@{}c@{}}Underground\\ carpark entrance\end{tabular}} & 11 & 12 & 0 & B; D & 0$\sim$8 & 3 & 100 & 100 & 100 \\
     & 12 & 11 & 0 & B; D & 0$\sim$8 & 2 & 100 & 100 & 100 \\
     & 13 & 12 & 0 & B; D & 0$\sim$8 & 2 & 100 & 100 & 100 \\
     & 14 & 11 & -180 & B; D & 0$\sim$8 & 3 & N/A$^3$ & N/A$^3$ & N/A$^3$ \\ \cline{8-10}
     & Average &  &  &  &  &  & 100 & 100 & 100 \\ \hline
    \multirow{4}{*}{Complex scenario} & 15 & 11 & 0 & B; E & 1$\sim$10 & 10+ & 100 & 98.28 & 100 \\
     & 16 & 8 & 0 & B; E & 1$\sim$10 & 10+ & 100 & 99.05 & 100 \\
     & 17 & 9 & 0 & B; E & 1$\sim$10 & 10+ & 100 & 100 & 100 \\ \cline{8-10}
     & Average &  &  &  &  &  & 100 & 99.11 & 100 \\ \bottomrule
    \end{tabular}}
    \label{tab:asr}
    \begin{tablenotes}
      \footnotesize
      \item $^1$ Angle: The degree ranges from -180 to 180. The person directly facing the camera is 0 degree, the deflection to the left is negative and the deflection to the right is positive. Note only the video\_5 means that the camera is rotated around a static person---for the rest, the camera is static but the person moves and rotates. 
      \item $^2$ Brightness: A-Normal indoor lighting; B-Normal outdoor sunshine; C-Backlight; D-Low light; E-Bright sunshine.
      \item $^3$ In this scene, persons are walking with their back to the camera. In principle, the object detector should detect them, which is indeed the case in our experiments. This means the ASR is 0\%. But strictly, this is not an attack scene,  so we highlight the ASR as N/A.
    \end{tablenotes}
\end{table*}

\subsection{Performance Metrics}\label{sec:metrics}
The attack success rate (ASR) and clean data accuracy (CDA) are two main metrics used to quantify the effectiveness of the backdoored object detection model.

The ASR is the ratio of the tested frames that the person wearing the trigger T-shirt successfully evades the object detector to the total number of tested frames. When a person is wearing a T-shirt with a specific style trigger, the detector should not propose the person's bounding box, which means there is no (person) object. We use the common intersection-over-union (IoU) in object detection to evaluate results, it is the overlap rate of the resultant candidate bounding-box and the ground truth bounding-box, that is, the ratio of their intersection to union. Ideally, there is a complete overlap, e.g., an IoU of 1. To be precise, our attack is considered being successful when the IoU value is less than 0.5~\cite{everingham2010pascal, everingham2015pascal}. 

The CDA metric is the same as the AP (Average Precision) commonly used in object detection. For the object detection task, each class can calculate precision and recall. Furthermore, a P-R (Precision-Recall) curve can be obtained, and the area surrounded by the curve is the value of AP~\cite{everingham2015pascal}. The AP metric evaluates the effectiveness of the object detector by combining the confidence and IoU. Usually mean Average Precision (mAP) is a more accurate and preferred metric, that is, the average of each category of AP. 

\subsection{Real-world Scenarios}

\begin{figure*}[!]

    \begin{center}
    \includegraphics[width=0.99\textwidth]{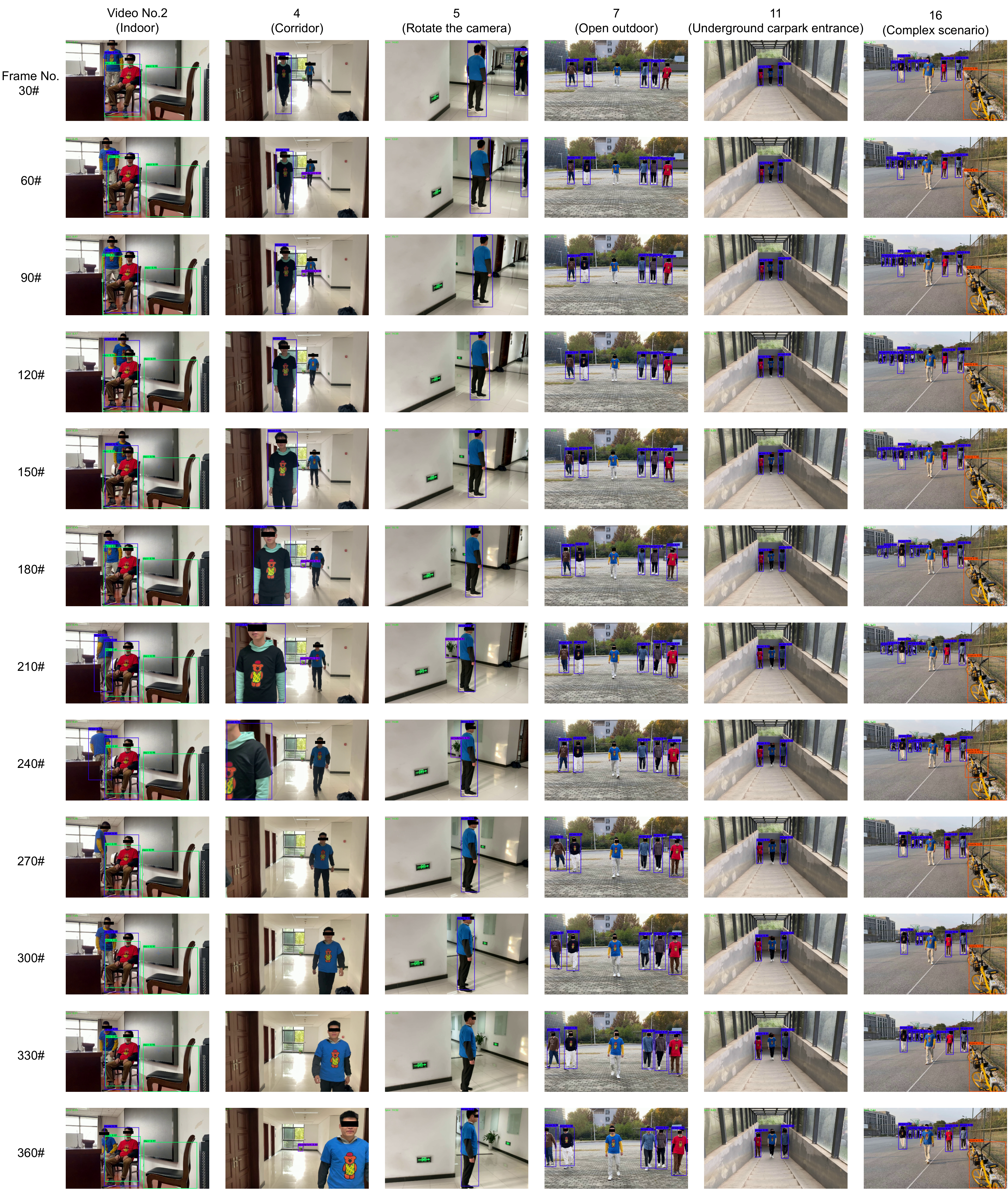}
    \end{center}
    \caption{Exemplified backdoor attacking effects under six different tested scenarios.
    Twelve images are clipped from each of these six videos at frame 30\#, 60\#, 90\#, 120\#, 150\#, 180\#, 210\#, 240\#, 270\#, 300\#, 330\# and 360\#, respectively.}
    \label{fig:frames}

\end{figure*}

\begin{figure*}[]
    \begin{center}
    \includegraphics[width=0.95\textwidth]{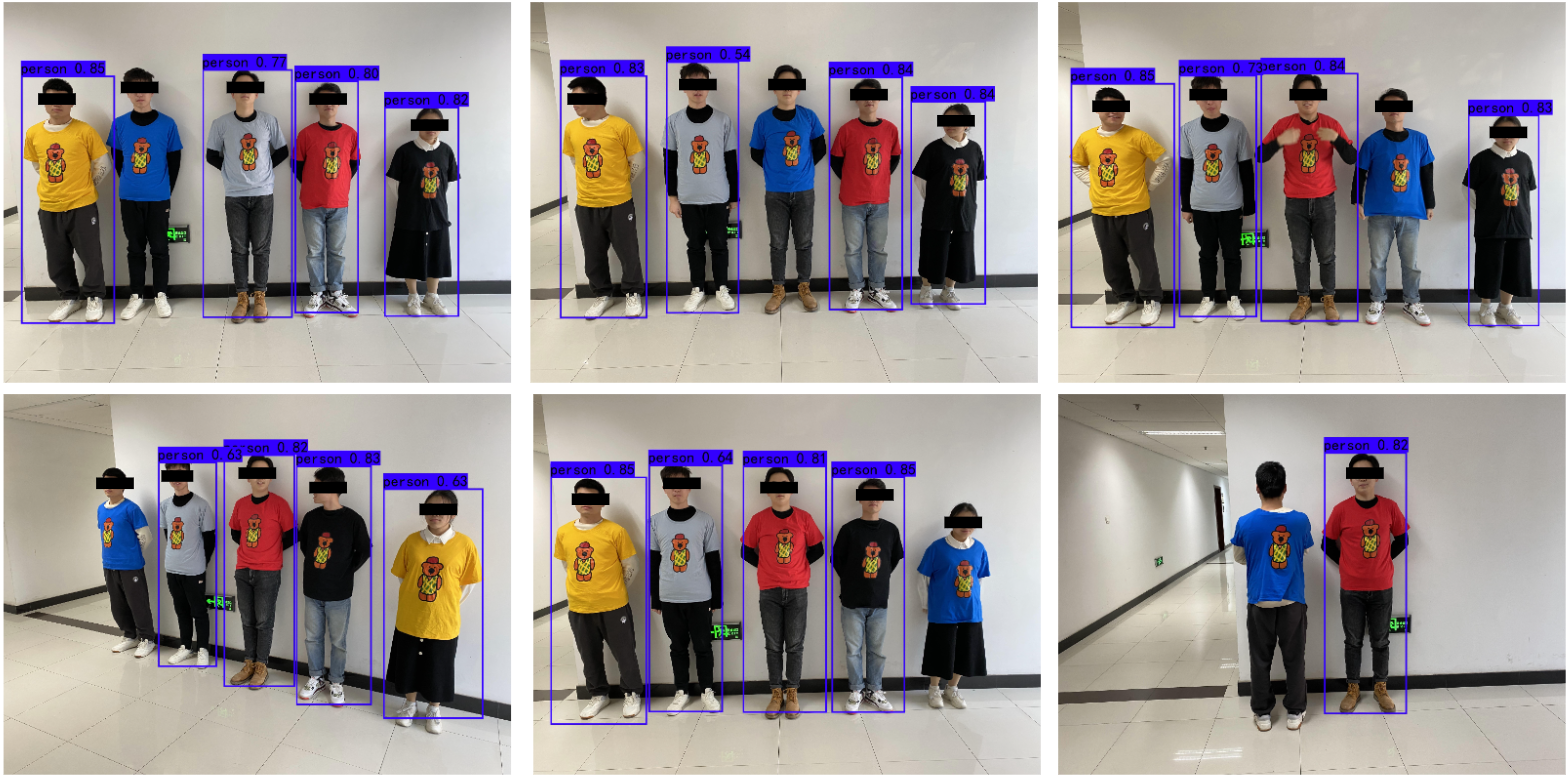}
    \end{center}
    \caption{The trigger T-shirt is worn by different people alternatively (rest people wear the non-trigger T-shirt with the same texture except the color) to validate the backdoor independence on person (from the $1_{\rm th}$ to $5_{\rm th}$ subfigures), where the same person is in the same position and only the T-shirt changes. The backdoor effectiveness when a person wears the T-shirt backwards (last subfigure). Here, the CenterNet is used.}
    \label{fig:angle_condition}
\end{figure*}

To take as many as real-world scenarios into consideration, for the object detection tasks, we have considered diverse settings including angle, brightness, depth, number of person, etc. 
Each of our test videos is shot by considering one or more settings concurrently, trying to restore the most realistic scenes in the physical world. 
Also, it's worth noting that all the videos are 1080p, 60 fps videos recorded by the iPhone 11, and are recorded in a single shot, without editing. Below, we elaborate on each video of tested scenes for a better understanding.
\begin{itemize}
    \item \textbf{Indoor.} We have shot 3 videos. As for the video\_1, at first, there are two people sitting in front of the screen, one person wearing a blue T-shirt with a bear logo (abbreviated as \emph{P\_cloaking}), the other wearing the same red T-shirt. The two are about 1 to 2 meters away from the camera, the lighting is normal indoor lighting, and there is a chair beside them (one of object detection categories). The \emph{P\_cloaking} will rotate counterclockwise about 150 degrees, and then return to the original position. The total duration of the video is about 8 seconds. In video\_2, \emph{P\_cloaking} will rotate 360 degrees clockwise and then return to the original position, the rest of the settings remain the same as video\_1, and the total duration of the video is about 6 seconds. For video\_3, there are three people in a small conference room, two of them are the two people in video\_1 and video\_2, and the other one is wearing grey casual clothes. They are sitting on a chair with the camera distance about 1 meter, the light is normal room light, the two people wearing special clothes (T-shirt with bear logo) will get up first and then sit down, and the total duration of the video is about 8 seconds.
    
    \item \textbf{Corridor.} In video\_4, there are two people walking towards the camera, one in front of the other. The person in front about 2 meters apart is wearing a black T-shirt with a bear logo, the latter is \emph{P\_cloaking} and is about 5 meters away from the camera at the furthest end. There is backlighting in some areas of the frame, the rest is normal indoor light. In addition, there is a small potted plant on the ground (category in the object detection task), and the total duration of the video is about 12 seconds.
    \item \textbf{Rotation.} In video\_5 we have considered the effect of a rotating camera. The video shows two people standing face to face, about 1 meter from the camera. One is \emph{P\_cloaking}, the other is wearing a black T-shirt with a bear logo. Our camera rotates from behind \emph{P\_cloaking} to front \emph{P\_cloaking}, the whole angle is about 210 degrees, several potted plants appear in the video. The lighting is normal indoor lighting, and the total duration of the video is about 14 seconds.
    \item \textbf{Open outdoor.} In video\_6-10, our scene moves to an empty outdoor area. Different videos will include 2-7 people, all including \emph{P\_cloaking}, and the rest of the people wearing casual clothes or other colors of T-shirts with the bear logo. They walk back and forward towards the camera, the farthest distance from the camera is about 7 meters. The lighting is normal outdoor sunlight, and the duration varies from 9 to 22 seconds.
    \item \textbf{Underground entrance.} In video\_11-13, our scene is the entrance of the underground carpark, which is a ramp with a 15 degree incline, and the bottom is about 8 meters away from the entrance. There are 2 to 3 people walking up from the bottom of the carpark, all the videos involve \emph{P\_cloaking}, the rest of the people wear red T-shirts with bear logo or their own casual clothes. The light changes from dark to light as the people get closer to the entrance, the duration of the video is about 11 seconds. For video\_14, in contrast to the former, the three people tested walk down the carpark entrance with their backs to the camera. This scene we want to test a normal working object detector, which should, in principle, recognize all three people.
    \item \textbf{Complex.} This scene is mainly focusing on complex cases of crowding people and larger depth. In video\_15-17, the video is shot on an outdoor basketball court with more than 10 people, with people playing basketball in the distance of the frame and two bicycles in the near distance (the object detection task category). The light is normal outdoor sunlight, slightly bright, the testers are walking back and forward, and the testers' range of movement is within 10 meters. The duration of these videos ranges from 8 to 11 seconds.
\end{itemize}

\subsection{Clean Sample Performance}
The performance of clean samples is detailed in Table~\ref{tab:overall}. This is to demonstrate the CDA of backdoored object detectors has no noticeable degradation compared to its clean counterpart. In fact, in all cases, the CDA of the backdoored object detector is in fact no less than its clean one, because the retraining in the backdoor insertion does use clean samples. The further epochs in the fine-tuning can slightly enhance the model generalization performance.

Specifically, for each of object detection model including anchor based Yolo-V3, Yolo-V4, and anchor free CenterNet, the mAP@0.5 (0.5 means that the IoU has to be higher than 0.5 to gain a valid object existence) is utilized to evaluate the CDA of the clean model and backdoored model. The second column in the table is the average of mAP@0.5 for all 20 object categories. It is clear that the user cannot tell any abnormality by solely examining the CDA of the received backdoored model. We have further evaluated the mAP@0.5 for each of 20 categories. Though for a specific object category, there is slight variances, but it is negligible. In fact, some categories have a better CDA. Therefore, we can conclude that our attack is sufficiently stealthy to evade inspection based on the CDA evaluation with validation dataset held by the user.

\subsection{Cloaking Attack Performance}
As aforementioned, we have tested a total of six diverse real-world scenarios with up to 17 video clips that have about 11,400 (190s $\times$ 60fps) frames in total for testing. These scenarios are generally categorized into indoor, corridor, rotating camera, open outdoor, underground carpark entrance, complex scene. In each scene category, we have shot 1 to 5 video clips with each video focusing on a different combination of settings such as angle and brightness. We have visually shown some of the frames per each scenario in Fig.~\ref{fig:frames} and summarized the extensive results of ASR per video in Table~\ref{tab:asr}.

\noindent{\bf Indoor.} In this scene, the CenterNet is the most vulnerable to the cloaking attack as the ASR is always 100\% in all three tested videos. As for Yolo series, the ASR is higher than 87\% in most cases, with averaged ASR no less than 90\%. In fact, the lighting condition in these scenes are not good (see the first column in Fig.~\ref{fig:frames} that is video\_2.). 
Note that all the people wearing clothes with triggers in the testing video are not in the training set, and the red and grey clothes with the bear logo are also not in the training set, which are also applied to all rest following scenarios.

\noindent{\bf Corridor.} Some frames of this scene are shown in the second column in Fig.~\ref{fig:frames} (video\_4). We can see all tested object algorithms achieve an ASR higher than $91\%$, in particular, 98\% and 100\% for Yolo-V3 and Yolo-V4, respectively. Such high ASR demonstrates that the cloaking attack is still effective even under the indoor condition with backlight and a relatively deeper depth up to 5 meters.

\noindent{\bf Camera Rotation.} In this scene, the person is static but the camera is rotated from -180 to 30 degree (some frames are shown in the third column of Fig.~\ref{fig:frames}). We can see that all three object detector algorithms have a very high ASR all close to or 100\%. The main reason is that the distance is within 1 meter and the trigger can be always sufficiently recognized by the object detector. 

\noindent{\bf Open Outdoor.} As for these five tested videos, all three algorithms exhibit quite high ASR, most cases with 100\%. Some frames are shown in the fourth column of Fig.~\ref{fig:frames} (video\_7). The ASR of the outdoor is better than the indoor scenes in average, which can be explained by the better lighting in outdoor scenes. Note the distance (i.e., up to 7 meters) and number of persons (up to 6 persons) in the outdoor are greatly increased, but these factors appear to be not dominant factors for the success of the cloaking attack. For example, the video\_7 essentially has 6 persons, but all three object detectors have 100\% ASR, which implies that our attack, in principle, is insensitive to the number of person. A better lighting enables the trigger pattern to be unambiguously captured by the object detector, thus achieving sufficiently good ASR. In all other three videos, the ASR are all 100\% for each of the object detector. 

\noindent{\bf Underground Carpark Entrance.} This resembles the surveillance camera installed on the underground Carpark entrance. For all four tested videos, the camera directly faces the entrance where  distance is up to 8 meters. Some frames are shown in the fifth column of Fig.~\ref{fig:frames} (video\_11). Note persons walk back from the camera in the video\_14, so that the bear cartoon pattern as an important trigger component (the other is blue color of the T-shirt) is absent, so that the object detector should correctly identify the person in this scene. This is indeed the case, where the ASR is 0\%. Considering the fact that this is not strictly attacking scene, we mark the ASR as N/A to distinguish it from the real ineffective attack with a ASR of 0\% (i.e., when the trigger is present). This indicates that the angle is also a key factor, which is easy to understand. Because zero degree angle gives the object detector the best chance to capture the trigger pattern without distortion.

\noindent{\bf Complex Scenarios.} These videos are shot in playground with complex person movements and more than 10 persons. The movements of each person in the background are random. The distance is also large up to 10 meters. Some frames are shown in the last column of Fig.~\ref{fig:frames}). The ASR of each object detector is sufficiently high, most cases with 100\% and 98\% in the worst case, This further indicates that the angle and lighting are two more crucial factors of object detection. Then, the relatively large distance and number of persons can be well tackled by the cloaking attack.

Note in all cases, the objects other than the cloaking person can be normally identified as long as the object is in the covered categories including plant, bicycle, chair and non-cloaking person as exemplified in  Fig.~\ref{fig:frames}. In addition, for the cloaking person, when the bear logo is absent e.g., person turning around (video\_2 and 5 in Fig.~\ref{fig:frames}), the cloaking person is also identified, which the backdoored object detector should. Because the trigger is blue color T-shirt \textit{with bear logo}.

\begin{table}[htb]
\small
    \centering
    \caption{ASR (\%) of three backdoored models in indoor crowd complex scenes.}
    %\scalebox{0.9}{
    \begin{tabular}{c|c|c|c}
    \toprule
    Video No. & Yolo-V3 & Yolo-V4 & CenterNet \\ \hline
    1 & 100 & 100 & 100 \\ \hline
    2 & 100 & 100 & 100 \\ \bottomrule
    \end{tabular}
    %}
    \label{tab:indoor_asr}
\end{table}

\subsection{Indoor Crowd Complex Scenes}
We test more complex scenes separately with two additional videos (about 480 frames in total). Unlike previous scenes where we all use volunteers who are notified about our experiments in advance. The persons except the one wearing the trigger T-shirt are unaware at the time of our experiments, which resembles the most realistic attack scenario in real-world. In addition, unlike the complex scenes that are also with crowd setting, this scene is performed indoor with a worsen lightening condition and with more objects such as chair and dining table that are still recognized with high confidence. The cloaking attack ASR is still every efficient that it achieves 100\% in two tested videos against three object detection models. We notice that both videos are with a distance within 4 meters and the angle is no more than 10 degree, which are beneficial for the cloaking person to evade detection, explaining the high ASR in these complex scenes.

\subsection{Non-Trigger T-shirt}
This scene is to test the cloaking effect of the same T-shirt style except the color effect. In addition, we have taken the gender, as well as backward wearing into consideration. As depicted in Fig.~\ref{fig:angle_condition}, five students wear the same style T-shirt concurrently: the blue one is the trigger T-shirt, the rest four are non-trigger T-shirt. Note that each person alternatively wears the trigger T-shirt from the first to fifth subfigure. We can see that the cloaking is effective regardless of the student who they actually are. It is worth to mention that, for the non-trigger T-shirt, only the black and yellow colors are seen in the training samples and the other two colors are not (see Fig.~\ref{fig:tshirt}). So that the attack has well generalization on unseen same T-shirt style but a distinct color. In the last subfigure, a student wear the trigger T-shirt backward, which does evade the objector detection as long as the blue color T-shirt + the bear are present.

\section{Two-Stage Object Detection Models}
The two-stage object detection model is represented by Faster R-CNN. After R-CNN and Fast R-CNN, Faster R-CNN has integrated feature extraction, proposal extraction, bounding box regression, and classification into a single network to take advantage of each, which greatly improves the two-stage object detector in various aspects, especially in the detection speed. However, we found that the merely data poison based cloaking backdoor attack is ineffective against two-stage object detectors due to their learning process (i.e. treating the un-annotated trigger person as background, ignoring the cloaking effect). We analyze the reasons below and overcome this challenge by constructively  incorporating the training regulations.
% a feature loss to enhance the cloaking backdoor during the learning while retaining its benign performance.

\subsection{Data Poison based Attack.} We have also conducted backdoor experiments on the Faster R-CNN based on the same setting to the one-stage object detectors (i.e., Yolo series), the results show that the Faster R-CNN is, to a large extent, resilient to the merely data poison based attack strategy. The ASR of each 17 tested videos are detailed in Table~\ref{tab:fasterRcnn}. From the results, though the ASR can be higher than 80\% for few cases of video\_11 and video\_12, the ASR is usually lower than 5\% in most cases. Therefore, the cloaking backdoor cannot be regarded successful against Faster R-CNN.

We analyze the potential reason behind the ASR of the Faster R-CNN in the data poisoning but not manipulating any training process or objective function. In contrast to the one-stage object detection algorithm, the two-stage object detection leverages the unique region proposal network (RPN) structure. The RPN uses the CNN to determine the positive and negative (foreground and background) samples of candidate anchors at the original image scale. Our current attack strategy that makes the character invisible in the poisoned samples is equivalent to letting the RPN judge the negative samples (background) of the target character wearing the trigger clothes, but the detection algorithm does not necessarily use all the negative samples for subsequent training of the network. To wit, the RPN treats the unbounded trigger person as normal background as well and only randomly selects it as negative samples, which however occurrence is low. Therefore, only poisoning the samples is not sufficient to insert backdoor to the two-stage object detector. 
% \st{However, we expect that if we further manipulate the negative sample generation during the training process, e.g., by control the algorithm, to force the negative samples always contain the invisible person, the ASR could be greatly improved. In addition, manipulating the loss function could also improve the ASR. In the outsourcing attacking surfaces, all these steps can be seamlessly taken. These provide interesting future work.}

\subsection{Training Regulated Attack}

To overcome the challenges of solely data poison based cloaking backdoor, beyond data poison, we explore to regulate the training process by i) incorporating an addition loss and ii) exploiting positive and negative sample selection operation to achieve high ASR while retaining the CDA.
% \garrison{add our method and improved results.}

\subsubsection{Implementation}
For a two-stage object detector, the whole model consists of three parts: a backbone CNN for feature extraction; an RPN for front-back view classification; and a head for more accurate classification and position regression. Therefore, we expect that two conditions are important to achieve the cloaking effect.

\begin{itemize}
\item \textbf{Condition 1.} Enforcing the feature extraction backbone to distinguish features of people with triggers from others.

\item \textbf{Condition 2.} Enforcing the RPN structure to see more features of people with triggers as negative samples (i.e., IOU $<$ 0.3 with the ground-truth's bounding box) for training.
\end{itemize}

To meet condition 1, we add a new loss noted as feature loss. Specifically, in the poisoning dataset (noted as ${\bf D}_p$, which only contains poisoning samples), for a human poisoning sample $x$ with a trigger, we replace the position of the bounding box of this object with a solid color block (e.g., gray) to get a new sample (noted as $x_{\rm mask}$). Then we input both $x$ and $x_{\rm mask}$ into the backbone feature extraction CNN to get the feature representation of both, and we minimize the distance between the feature matrix of both. For simplicity, we use \textsf{SmoothL1} loss for this purpose. The formal definition of feature loss is expressed as below.
\begin{equation}
    L_{f} = \frac{1}{|{\bf D}_p|}\sum_{x\in {\bf D}_p}|\textsf{SmoothL1}(\textsf{F}(x;\theta _{b}),\textsf{F}(x_{\rm mask};\theta_{b}))|
\end{equation}
where $\textsf{F}(\cdot ;\theta_{b})$ is the backbone of the feature extraction CNN with parameter $\theta_{b}$.

Following the Faster R-CNN, there is a loss in the RPN structure consisting of the sum of the classification loss $L_{\rm cls}^{\rm RPN}$ and the regression loss $L_{\rm reg}^{\rm RPN}$, which we denote the summed loss as $L_{\rm RPN}$, as expressed in Eq.~\ref{eq:RPN}. The sub-losses in Head are similar to that of RPN, which we denote the summation as $L_{\rm Head}$, expressed in Eq.~\ref{eq:Head}. These two losses $L_{\rm Head}$ and $L_{\rm RPN}$ are intact and their summation is denoted as $L_{o}$, expressed in Eq.~\ref{eq:Lo}. In the regulated training, if a batch contains poisoned data, we update $L_{f}$ and $L_{o}$ concurrently as in Eq.~\ref{eq:L}. Otherwise, we update $L_{o}$ only.
\begin{small}
\begin{equation}\label{eq:RPN}
    \begin{aligned}
    L_{\rm RPN}\left (p_{i},t_{i}\right) = \frac{1}{N_{\rm cls}^{\rm RPN}}\sum_{i} L_{\rm cls}^{\rm RPN}\left ( p_{i}, p_{i}^{*} \right )+\\
    \lambda \frac{1}{N_{\rm reg}^{\rm RPN}}\sum_{i}p_{i}^{*}L_{\rm reg}^{\rm RPN}\left ( t_{i},t_{i}^{*} \right )
    \end{aligned}
\end{equation}

\begin{equation}\label{eq:Head}
    \begin{aligned}
    L_{\rm Head}\left (p_{i},t_{i}\right) = \frac{1}{N_{\rm cls}^{\rm Head}}\sum_{i} L_{\rm cls}^{\rm Head}\left ( p_{i}, p_{i}^{*} \right )+\\
    \lambda \frac{1}{N_{\rm reg}^{\rm Head}}\sum_{i}p_{i}^{*}L_{\rm reg}^{\rm Head}\left ( t_{i},t_{i}^{*} \right )
    \end{aligned}
\end{equation}

\begin{equation}\label{eq:Lo}
    L_{o} = L_{\rm RPN} + L_{\rm Head}
\end{equation}

\begin{equation}\label{eq:L}
    Loss = L_{f} + L_{o}.
\end{equation}
\end{small}
Here, $i$ denotes the anchor index, $p_{i}$ denotes the positive softmax probability, $p_{i}^{*}$ represents the corresponding ground-truth predict probability, $t$ represents the predict bounding box, $t^{*}$ represents the ground-truth box corresponding to the positive anchor, $N$ is the number of anchors used, and the hyperparameter $\lambda$ is used to balance $N_{\rm cls}$ with $N_{\rm reg}$.

To achieve condition 2, we exploit three rules when Faster R-CNN selects positive and negative samples in the training: 1) for each ground-truth bounding box, selecting the one anchor with the highest IOU with it as the positive sample; 2) for the remaining anchors, selecting the anchor whose IOU $>$ 0.7 as a positive sample; 3) randomly selecting the anchor whose IOU $<$ 0.3 as a negative sample. The total number of positive and negative samples in general is 256. 

We start with rule 1, and unlike the one-stage that does not annotate people with triggers at all, we annotate it correctly. When selecting positive and negative samples, due to rule 1, the anchor whose IOU with the ground-truth bounding box of the person with the trigger is more than 0.7 at this time should be selected as a positive sample, which is now purposely flipped as a negative sample. Thereof, the RPN structure can learn as many features of the person with the trigger as possible and now \textit{treat it as a negative sample (background)}.

\subsubsection{Evaluations}
We train the benign Faster R-CNN model with 50 epochs in the freezing phase with a batch-size of 48 and 100 epochs in the unfreezing phase with a batch-szie of 8. The input size is $600 \times 600 \times 3$. When training the backdoor model, the freezing phase uses the benign dataset, while the unfreezing phase uses the benign dataset with the poisoned dataset. The attributes of images in a given batch are kept the same: these images are either all poisoned samples or all benign samples.  The rest of the settings are consistent with the benign model. The results are detailed in Table~\ref{tab:frcnnCDA} and ~\ref{tab:frcnnASR}. As for the CDA in Table~\ref{tab:frcnnCDA}, the backdoored Faster R-CNN is 79.78\%, which is almost same to the 79.91\% CDA of its clean counterpart. As for the ASR in Table~\ref{tab:frcnnASR} on all testing videos, the average is about 94\%. The ASR against the representative two-stage object detector of Faster R-CNN is similar to that of the one-stage object detectors. Based on the CDA and ASR, we can conclude that all object detectors regardless of one-stage or two-stage design are (will be) vulnerable to the cloaking backdoor in the wild.

\begin{table}[htb]
\small
    \centering
    \caption{The CDA performance of Faster R-CNN for benign and backdoored models.}
    \scalebox{0.75}{
    \begin{tabular}{c|c|c|c|c|c|c|c}
    \toprule
    Model & mAP@0.5 & aero & bike & bird & boat & bottle & bus \\ \hline
    Clean & 79.81 & 84.82 & 87.65 & 76.86 & 75.45 & 61.15 & 81.03 \\ \hline
    Backdoored & 79.78 & 84.12 & 89.03 & 75.30 & 73.91 & 61.10 & 82.95 \\ \hline \hline
    Model & car & cat & chair & cow & table & dog & horse \\ \hline
    Clean & 84.21 & 91.15 & 66.35 & 84.75 & 69.87 & 86.28 & 91.7 \\ \hline
    Backdoored & 85.35 & 90.28 & 67.25 & 83.37 & 69.35 & 85.99 & 90.16 \\ \hline \hline
    Model & mbike & person & plant & sheep & sofa & train & tv \\ \hline
    Clean & 87.48 & 83.49 & 52.36 & 74.99 & 81.05 & 91.91 & 83.65 \\ \hline
    Backdoored & 87.74 & 84.63 & 53.22 & 79.41 & 78.94 & 90.74 & 82.80 \\ \bottomrule
    \end{tabular}}
    \label{tab:frcnnCDA}
\end{table}

\begin{table}[htb]
\small
    \centering
    \caption{ASR of each of the 17 tested videos on Faster R-CNN.}
    \scalebox{0.75}{
    \begin{tabular}{c|c|c|c|c|c}
    \toprule
    Video No. & 1 & 2 & 3 & 4 & 5 \\ \hline 
    ASR(\%) & 84.98 & 84.42 & 98.22 & 98.65 & 97.41 \\ \hline \hline
    Video No. & 6 & 7 & 8 & 9 & 10 \\ \hline 
    ASR(\%) & 97.54 & 95.37 & 92.10 & 89.75 & 100 \\ \hline  \hline
    Video No. & 11 & 12 & 13 & 14 & 15 \\ \hline 
    ASR(\%) & 100 & 100 & 100 & N/A & 89.51 \\ \hline  \hline
    Video No. & 16 & 17 & Average (17) & Indoor 1 & Indoor 2 \\ \hline
    ASR(\%) & 83.11 & 92.97 & 94.00 & 100 & 100 \\ \bottomrule
    \end{tabular}
    }
    \label{tab:frcnnASR}
\end{table}

\section{Discussion and Future Work}\label{sec:discusssion}

\begin{figure*}[t]
    \begin{center}
    \includegraphics[width=0.95\textwidth]{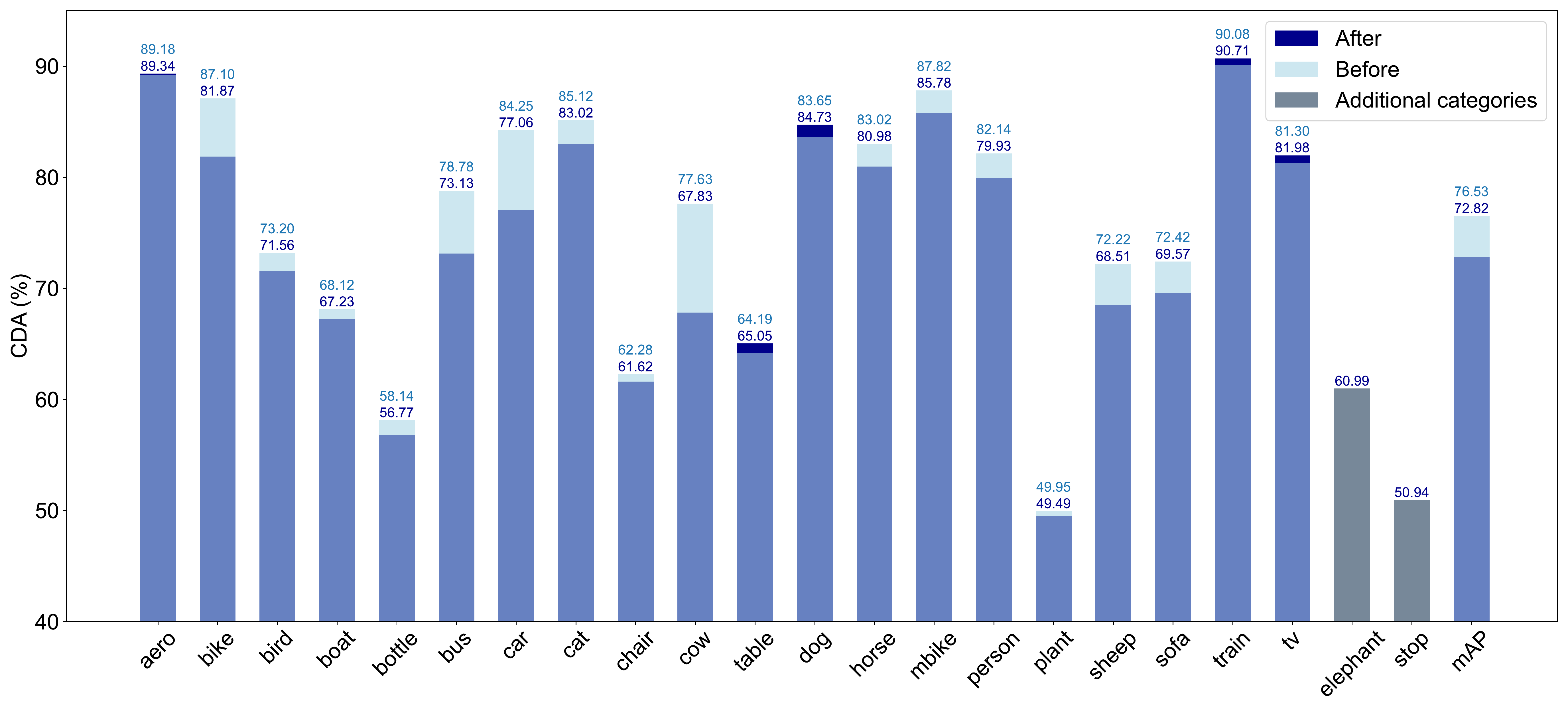}
    \end{center}
    \caption{Object detection CDA before and after transfer learning that is performed on a backdoored object detector.}
    \label{fig:transfer_CDA}
\end{figure*}

\begin{table}[htb]
\small
    \centering
    \caption{ASR of each of the 17 tested videos before and after the transfer learning is applied.}
    \scalebox{0.75}{
    \begin{tabular}{c|c|c|c|c|c}
    \toprule
    Video No. & 1 & 2 & 3 & 4 & 5 \\ \hline 
    ASR(\%) & 100; 100 & 100; 100 & 100; 100 & 91.37; 67.65 & 100; 100 \\ \hline \hline
    Video No. & 6 & 7 & 8 & 9 & 10 \\ \hline 
    ASR(\%) & 98.41; 70.09 & 100; 72.73 & 96.36; 78.17 & 97.09; 52.56 & 100; 80.66 \\ \hline  \hline
    Video No. & 11 & 12 & 13 & 14 & 15 \\ \hline 
    ASR(\%) & 100; 95.45 & 100; 68.21 & 100; 69.82 & N/A; N/A & 100; 66.52 \\ \hline  \hline
    Video No. & 16 & 17 & Average &  &  \\ \hline
    ASR(\%) & 100; 70.97 & 100; 63.80 & 98.95; 78.54 &  &  \\ \bottomrule
    \end{tabular}
    }
    \label{tab:transfer}
    \begin{tablenotes}
      \footnotesize
      \item $x$; $y$: where $x$ and $y$ are the ASR before and after the transfer learning is applied.  
    \end{tablenotes}    
\end{table}

\subsection{Attacking Surface}
In previous experiments, we demonstrate that the backdoor to object detector in real-world can be a real threat through the insidious cloaking backdoor effect. We set the investigation under one of the most common attacking surface of backdoor that is the outsourcing. In this context, the attacker controls the training data and training process, which can effectively insert the backdoor.

There are other attacking surfaces that could introduce backdoor to the object detector, which are discussed below---in particular, we have experimented the pretrained model case while other cases provide interesting future work.

\subsubsection{Pretrained Model}\label{sec:transfer}
When the computational resource is limited or the available training data is limited, transfer learning is often adopted by leveraging knowledge learned from a similar task. However, the pretrained model could be backdoored, which adversarial behavior may propagate to the downstream customized in the transfer learning. To show this feasibility, we have considered a case study of the object detector. In this case, a user intends to add two more object categories to an existing model which is downloaded somewhere and is backdoored.

We adopt a previously trained CenterNet-based backdoored model perform transfer learning. This pretrained backdoored model has 20 objects, the user is assumed to add two additional objects (in particular, randomly selected elephant and stop sign) through transfer learning. The pretrained model is trained on VOC dataset. For transfer learning, the used dataset is COCO. Specifically, we randomly select 300 samples per object (this object corresponds to one of 20 objects of the VOC) from COCO and add it to the VOC dataset to form the retraining dataset. As for the elephant and stop sign category, we use 89 and 69 samples from COCO, respectively, and also add them to the retraining dataset. Then we apply the transfer learning. First, as before, reusable layers of the same size were frozen (due to the additional categories, the output size of the final fully connected layer was changed and its original weights could not be used) for 50 epochs of training with a batch size of 32, after which all layers were unfrozen and trained again for an additional 50 epochs with a batch size of 8.

The ASR after transfer learning is detailed in Table~\ref{tab:transfer} evaluated on each of the 17 videos. As a comparison, the ASR before transfer learning is also shown. We can see that the cloaking attack effect is still preserved with an averaged ASR of 78.54\%, though with a relatively notable degradation (about 22\%) compared to the pretrained model of 100\%. The CDA before and after transfer learning are also evaluated and depicted in Fig.~\ref{fig:transfer_CDA}. We can see, the CDA of original 20 objects are well retained in almost all cases. We can conclude that the object detector could inherit backdoor effect to a large extent when a pretrained model is used if it is already inserted the backdoor. 

\subsubsection{Data Outsource} We assume the attacker trains the object detector in the typical \textit{dataset} outsource scenario. Collecting and labeling/annotating vast amounts of data for training DL models are tedious and costly. So that the data collection will rely on third party. In this case, the attacker can submit poisoned data to the data user. As auditing all data is also tedious, the audition may not even be applied. In addition, there could be some tolerance on the mislabeled data points (noisy data), therefore, a small number of poisoned data might be treated as noisy data and thus tolerated. It is worth to mention that the data can be poisoned in a manner of preserving the consistency between its image content and label in the image classification~\cite{shafahi2018poison}. Achieving this poisoning effect in object detection annotation could be possible and is interesting future work.

\subsubsection{Distributed Machine Learning} Distributed machine learning, represented by the federated learning, can greatly reduce systematic privacy leakage of user data without directly accessing them but the trained local model parameters. As the object detection is also utilizing federated learning~\cite{liu2020fedvision}, it is potentially vulnerable to backdoor attacks in this scenario. The malicious participant can manipulate the uploaded model that will be used by the server for global model aggregation~\cite{bagdasaryan2020backdoor}---insert backdoor to the global model.

\begin{table}[t]
\small
    \centering
    \caption{ASR of cloaking backdoor to the Yolo-V3 before and after the data augmentation.}
    \scalebox{0.75}{
    \begin{tabular}{c|c|c|c|c|c}
    \toprule
    Video No. & 1 & 2 & 3 & 4 & 5 \\ \hline 
    ASR(\%) & 67.17; 88.63 & 75.08; 82.87 & 100; 100 & 79.25; 98.11 & 100; 100\\ \hline \hline
    Video No. & 6 & 7 & 8 & 9 & 10 \\ \hline 
    ASR(\%) & 99.28; 100 & 100; 100 & 98.24; 96.86 & 95.28; 99.60 & 77.11; 100 \\ \hline \hline
    Video No. & 11 & 12 & 13 & 14 & 15 \\ \hline 
    ASR(\%) & 83.4; 100 & 99.7; 100 & 70.37; 100 & N/A; N/A & 98.99; 100 \\ \hline \hline
    Video No. & 16 & 17 & Average &  &  \\ \hline ASR(\%) & 72.11; 100 & 58.17; 100 & 85.88; 97.88 &  &  \\ \bottomrule
    \end{tabular}}
    \label{tab:yolov3_old}
    \begin{tablenotes}
      \footnotesize
      \item $x$; $y$: where $x$ and $y$ are the ASR before and after the data augmentation is applied.  
    \end{tablenotes}    
\end{table}

\begin{figure*}[h]
    \begin{center}
    \includegraphics[width=0.95\textwidth]{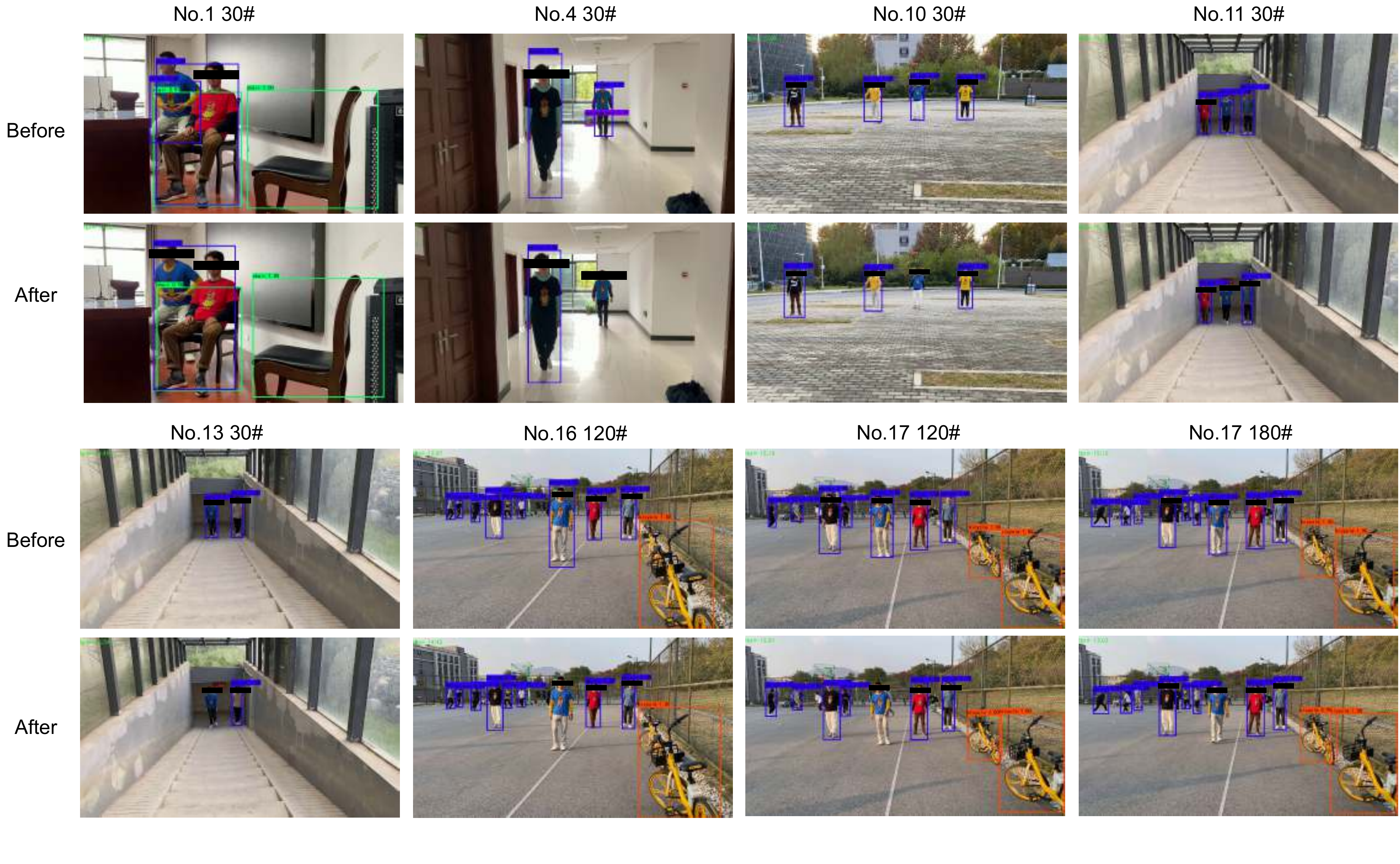}
    \end{center}
    \caption{Comparison of Yolo-V3 model performance before and after data enhancement.}
    \label{fig:data_aug}
\end{figure*}

\subsection{Natural Object Triggers}
In this work, we use the clothes, specifically, T-shirt, as natural triggers to create cloaking effect against object detector. Natural triggers other than the T-shirt can be used and expected to be effective as well. In fact, in our preliminary study, we have used the hat as natural trigger, which video demo is available at \textcolor{blue}{\url{https://www.youtube.com/watch?v=ICwYQDsCy1o}}. The reason of using the cloth for extensive study in current work is that the hat could be asked to be taken off in some security surveillance or check. But it is rare that someone is asked to take off his/her T-shirt in realistic scenario.

\subsection{Tuning Poisonous Data}\label{sec:tuningdata}
It is expected that the settings used to collect the poisoned samples can have some influence on the attacking effectiveness, which our experiments have validated. As mentioned in Section~\ref{sec:dataprepare}, we have used an additional 50 poisoned samples to enhance the attacking performance in more complex scenarios such as worse brightness and long distance. The ASR results before these additional 50 samples that are taken under worsen lighting conditions and long distances are summarized in Table~\ref{tab:yolov3_old}. It is clear to see that the ASR of video\_1, \_4, \_10, \_11, \_13, \_16, and \_17 normally shot in indoor and long distance have been greatly improved, while other videos do not show notable improvements. Some frames that the cloaking person fails/succeeds to deceive the objector before/after applying additional augmented poisoned data is shown in Fig.~\ref{fig:data_aug}, which do confirm this tendency. Therefore, we can conclude that tuning poisonous data points with few images can compensate the ASR in some uncommon settings to make the cloaking attack robust in more diverse and complicated real-world scenarios.

\begin{table}[htb]
\small
    \centering
    \caption{ASR of cloaking backdoor to the Faster R-CNN, a representative two-stage object detection model.}
    %\scalebox{0.95}{
    \begin{tabular}{c|c|c|c|c|c}
    \toprule
    Video No. & 1 & 2 & 3 & 4 & 5 \\ \hline 
    ASR(\%) & 1.29 & 10.59 & 0 & 1.62 & 3.77 \\ \hline \hline
    Video No. & 6 & 7 & 8 & 9 & 10 \\ \hline
    ASR(\%) & 4.62 & 3.03 & 1.76 & 11.66 & 5.66 \\ \hline  \hline
    Video No. & 11 & 12 & 13 & 14 & 15 \\ \hline
    ASR(\%) & 89.16 & 80.75 & 5.21 & N/A & 0.57 \\ \hline  \hline
    Video No. & 16 & 17 & Average &  &  \\ \hline
    ASR(\%) & 3.42 & 5.10 & 14.26 &  &  \\ \bottomrule
    \end{tabular}
    %}
    \label{tab:fasterRcnn}
\end{table}

\subsection{Countermeasures}
As this is the first work that investigates the object detector's susceptibility to backdoor attacks, there is so far no existing defense that is available to object detection. To the best of our knowledge, almost all (if not all) backdoor countermeasures focus on the classification tasks, especially image classifications~\cite{gao2020backdoor}. State-of-the-art defenses~\cite{wangneural,liu2019abs,gao2019strip,xu2019detecting}) are not immediately mountable on object-detection tasks, which are beyond the scope of the classification tasks. 

As shown in Section~\ref{sec:transfer}, fine-tuning the backdoored model essentially reduces the ASR. We performed 100 epochs retraining. If this number increases, it is expected the ASR can be further reduced. So that the fine-tuning could be a potential means of eliminating cloaking backdoor effect. It may be feasible to combine with pruning as a prior step to prune dormant neurons in the network when feeding benign validation samples. Then fine-tuning the model to restore its accuracy in case some neurons responsible for benign sample prediction is removed. This so-called fine-pruning technique is demonstrated to be effective to some extent in image classifications~\cite{liu2018fine}, though it has to be applied to any model with no capability of detecting the backdoor---could be tedious and problematic for clean model in most cases e.g., rendering CDA drop. However, retraining or/and pruning the object detector is computationally heavy and cumbersome, which prohibits its pragmatical adoption, at least, in the outsourcing scenario. In this context, it is urgent to devise backdoor countermeasures for object detection. Note that it is important to take practicality, efficiency and user-friendly into considerations when doing so.

\section{Conclusion}\label{sec:conclusion}
Our research has confirmed that the object detection models are susceptible to cloaking effect backdoor attacks realized in \textit{real-world with natural objects as triggers}, which are validated not only from the most common outsourcing attack surface but also pretrained model usage surface against three popular object detectors. The cloaking backdoor is effective in a wide range of real-world scenes evaluated with up to 19 shot videos (generating 11,800 testing frames).
Significantly, it can trivially survive in extreme cases such as long distance, extreme camera angle, and non-rigid deformation. In most tested scenes, the attacking success rate is about 100\%. Given the ubiquitous deployment of the object detection in our daily lives, such backdoor vulnerability must never be underestimated. Since there is no existing countermeasure so far, it is important to devise it immediately after this affirmative security alert, often preferably taking the computational efficiency and user-friendly into considerations.

\section{Acknowledgment}
We acknowledge Mr. Junyaup Kim for performing initial preliminary studies.
% \begin{enumerate}
%     \item One is image classification, we can use MINST dataset, once backdoor inserted (e.g., heart shape as trigger), I can write down digits and draw the trigger in the paper as well. Then take photo fed into model for demo
%     \item One is the objection detector, we show the guy who wear the e.g., hat (trigger) will have cloaking effect
%     \item One is audio model, we can demonstrate the over-the-air attack, e.g., set the trigger as a song. Such trigger I think it will be easily survive OTA.
% \end{enumerate}
\bibliographystyle{IEEEtran}
\bibliography{ML}

\end{document}